\documentclass[10pt]{article}
\usepackage{graphicx}
\usepackage{amssymb,amsthm,amsmath}
\usepackage{xcolor,paralist,hyperref,titlesec,fancyhdr,etoolbox}
\usepackage{authblk} 

\hypersetup{
    colorlinks=true,
    linkcolor=blue,      
    citecolor=blue,      
    filecolor=blue,      
    urlcolor=blue        
}

\title{If our aim is to build morality into an artificial agent, how might we begin to go about doing so?}
\author[1]{Reneira Seeamber}
\author[1]{Cosmin Badea}
\affil[1]{Department of Computing, Imperial College London, London, SW7 2BX, United Kingdom}
\date{\today}

\begin{document}
\maketitle

\begin{abstract}\looseness-1As Artificial Intelligence (AI) becomes pervasive in most fields, from healthcare to autonomous driving, it is essential that we find successful ways of building morality into our machines, especially for decision-making. However, the question of what it means to be moral is still debated, particularly in the context of AI. In this paper, we highlight the different aspects that should be considered when building moral agents, including the most relevant moral paradigms and challenges. We also discuss the top-down and bottom-up approaches to design and the role of emotion and sentience in morality. We then propose solutions including a hybrid approach to design and a hierarchical approach to combining moral paradigms. We emphasize how governance and policy are becoming ever more critical in AI Ethics and in ensuring that the tasks we set for moral agents are attainable, that ethical behavior is achieved, and that we obtain good AI.
\end{abstract}

If our aim is to build morality into an artificial agent, or machine, how might we begin to go about doing so? This is a crucial question that is of particular relevance at present, in the context of the flurry of recent activity in the field, including recent but already famous applications such as ChatGPT and Bard. 

Let us begin by defining key terms in this question, namely ‘morality’ and ‘artificial agent’. Morality can be defined as the differentiation of intentions, decisions, and actions into those that are good (or right) and those that are bad (or wrong) \textsuperscript{\cite{khatibi_morality_2016}}. An artificial agent, or machine, we will consider to be a system that can intelligently perform certain tasks by making decisions.

Deciding whether or not we should build morality into artificial intelligence (AI) is determined by many factors, including whether we want the AI to act as a moral agent or adviser and whether we want to build autonomy into it. Going down the latter route brings with it an obvious need for morality to be implemented, but the fundamentals of what it is to be 'good' have long been debated and no clear consensus has been reached\textsuperscript{\cite{badea_have_2022}}. 

To begin addressing this dilemma, we can look at the three main philosophical theories of morality in AI Ethics: virtue ethics, utilitarianism, and deontology\textsuperscript{\cite{badea_have_2022}}. Deontology emphasizes the imperative moral value of certain actions or rules, rather than their consequences. Virtue ethics, on the other hand, focuses on the character of the individual who is acting, rather than on specific rules or actions. It emphasizes the importance of developing virtuous traits, such as honesty and courage and developing a virtuous character, which allows individuals to make morally good choices. Utilitarianism is a consequentialist theory that focuses on actions that maximize utility, commonly measured as happiness or pleasure, for the greatest number of people \textsuperscript{\cite{badea_have_2022}}.

Existing research has investigated different approaches to developing good AI and important obstacles and potential solutions have been identified. For instance, the \textit{Interpretation Problem}, introduced by Badea and Artus (2022)\textsuperscript{\cite{badea_morality_2022}} is the idea that any symbolic representation we might use for AI necessarily gives rise to a plethora of possible interpretations, and thus ambiguity and mistakes by the AI. There are ways to mitigate its effects, such as building character into the agents, thus creating virtuous agents by using values as a tether between what the AI understands and what we mean to communicate to it \textsuperscript{\cite{badea_morality_2022}}. There are also more practical questions, such as those around what role agents would play in a clinical context, with proposals including them acting alongside the human experts, perhaps relying on such experts as moral exemplars. \textsuperscript{\cite{hindocha_moral_2022}}. Additionally, there are arguments around the use of particular moral paradigms (such as utilitarianism, deontology and virtue ethics) in building ethical AI and resolving contemporary moral dilemmas like breaking bad news to a patient \textsuperscript{\cite{post_breaking_2022}} or antimicrobial resistance and use  \textsuperscript{\cite{bolton_developing_2022}}. Furthermore, the variety of potential moral paradigms and lack of agreement has also led to a perceived necessity for a moral paradigm-agnostic decision-making framework, which Badea (2022) proposes as \textit{MARS} \textsuperscript{\cite{badea_have_2022}}. However, building decision-making agents is itself a difficult process, and deciding on how to do it, or Meta-decision-making for AI \textsuperscript{\cite{badea_establishing_2022}}, is a new and very promising field of research, which we aim to contribute to here. An ontology for this process has been proposed before, based on three steps, in Badea and Gilpin (2022).
In this paper, we continue previous attempts to identify methods and challenges for building good AI, and to propose solutions that overcome these challenges. With this work, we examine canonical approaches to AI to provide a broader perspective on implementing morality. Having identified many different challenges and potential ways of dealing with them, we found ourselves in the position of wanting to take a step back and focus on the techniques of building ethical AI. Thus, we examined existing methodology from the literature and identified top-down and bottom-up approaches from classical AI \textsuperscript{\cite{abel_reinforcement_2016}}. Having identified these canonical approaches, we aimed to apply them to the task of mitigating the issues discussed above. Hence we set out with the goal of coming up with and categorizing such methods, and continuing this line of thought in attempting to examine the potential technical approaches to implementing the suggestions above and constructing working ethical AI. 

\section{THE TOP-DOWN APPROACH ALONE MIGHT BE INSUFFICIENT}

An important aspect to consider in designing AI is the influence of culture and how it affects morals. Massachusetts Institute of Technology (MIT) created the Moral Machine, an online simulation where the public can decide for themselves which is the ‘right’ action to take given variations of the famous ‘Trolley dilemma’ \textsuperscript{\cite{awad_moral_2018}}. The dilemma originally involved choosing between allowing a group of five people to die or choosing to pull a lever
and sacrificing one person instead, while the Moral Machine creates variations on this with different demographics
in each of the two groups. Global data from the MIT study revealed variations in how people respond to such moral dilemmas, given their demographic. For example, people in Western countries were more likely to favor saving younger people than the elderly compared with people in Eastern countries. This casts doubt on the possibility of using a top-down approach alone because implementing such approaches (utilitarianism, deontology) might be nigh impossible, given that there may never be a global consensus between humans on developing a universal set of ‘rules’ or ‘duties’.

Noothigattu et al. (2018) proposed a voting-based system, whereby data from the moral machine is collected and used to train multiple models to learn societal preference for alternative outcomes across multiple ethical dilemmas. These models are then aggregated to form a model that considers the collective outcome of all voters, thus satisfying the utilitarian approach. This method is limited in that self-reported preferences have been shown to vary from real-life decisions \textsuperscript{\cite{noothigattu_voting-based_2018}}. Additionally, it may be argued that few AI developers would have to deal with ethical dilemmas, such as the Trolley problem, in their AI application and the majority of humans have not or will not have to make such a decision. Instead, we should be looking towards more realistic dilemmas, such as social media advertising and whether it could lead to accessibility to violence or compulsive behavior formation. It is up for debate whether a set of deducible principles could be created and whether a model needs more than simply a set of internal coherence constraints, as it could easily satisfy constraints but still produce undesirable outcomes. In reality, multiple goals and rules come into conflict with each other, producing complexity. Developing rules that a machine will understand in the way we desire is another issue in itself, which may stem from us not knowing explicitly what it is we want them to understand  \textsuperscript{\cite{badea_morality_2022}}.

In Plato’s dialogue Euthyphro, Socrates asked Euthyphro, “Is the pious loved by the Gods because it is pious or is it pious because it is loved by the Gods?”. This is proving very relevant to AI morality as we, humans, may begin to see ourselves as the Gods, or creators, of AI. Thus, is it what we (subjectively) believe morality to be that AI should consider as moral, or should we leave it to the AI to determine what is moral for itself, potentially then leading us to reassess what we see as moral? For instance, the aggregation of societal preferences has the potential to produce a morally better system than that of any individual alone, and may even enable us to then identify general principles that underlie our own decision-making \textsuperscript{\cite{badea_establishing_2022}}.

\section{EMOTION, SENTIENCE AND MORALITY}

It has been debated, especially following Jeremy Bentham's and John Stuart Mill's formulation of ‘the greater good for the greatest number’, whether the stoic characteristic of machines limits their ability to carry out moral decision-making because of their incapacity to care. The dual-process theory of moral judgment, which proposes the idea of dichotomous thinking, with a fast, instinctive and emotional process driving us alongside a slower and more logical, reason-guided process \textsuperscript{\cite{wallach_moral_2009}}, has led to the question: Does this emotional capacity of humans lead us to moral subjectivity? Or does it allow us to decide between what is wrong or right? Thinking of humans as having two different systems may play an important role in developing AI machines that have both reason and the capacity to care. ‘Cognitive emotion’ could be built into machines, whereby one allows for weights to be added to, for example, ‘human life’, thereby producing somewhat similar caring systems for human life \textsuperscript{\cite{wallach_moral_2009}}.

However, the innate stoicism of machines may be advantageous in reducing the moral biases we have as humans, one of the limiting factors as to why it is hard to come to a consensus \textsuperscript{\cite{wallach_moral_2009}}. The idea of ‘embodiment’ and whether artificial moral decision-making should be extended with emotional and sensory components is interesting. However, various issues come to light when extensive sensor systems are involved. For example, these are more vulnerable to attack, and people may become worried about privacy issues.

Whether or not a sentient agent is achievable remains to be explored. The symbiosis between technology and humans will become ever more important, and similar to a jury determining someone’s innocence, or a group of doctors discussing how to proceed with the most complex medical problems, there are simply certain decisions that require several inputs before coming to a conclusion \textsuperscript{\cite{hindocha_moral_2022}}. It is therefore imperative that any system taking on responsibilities that would otherwise be discussed by a group of people should have a notification state. This state would ensure that a board of human overseers intervene and is involved in the final decision when conflicts arise. 

\section{PROPOSING A HYBRID APPROACH}

\begin{figure}
    \centering
    \includegraphics[scale=0.7]{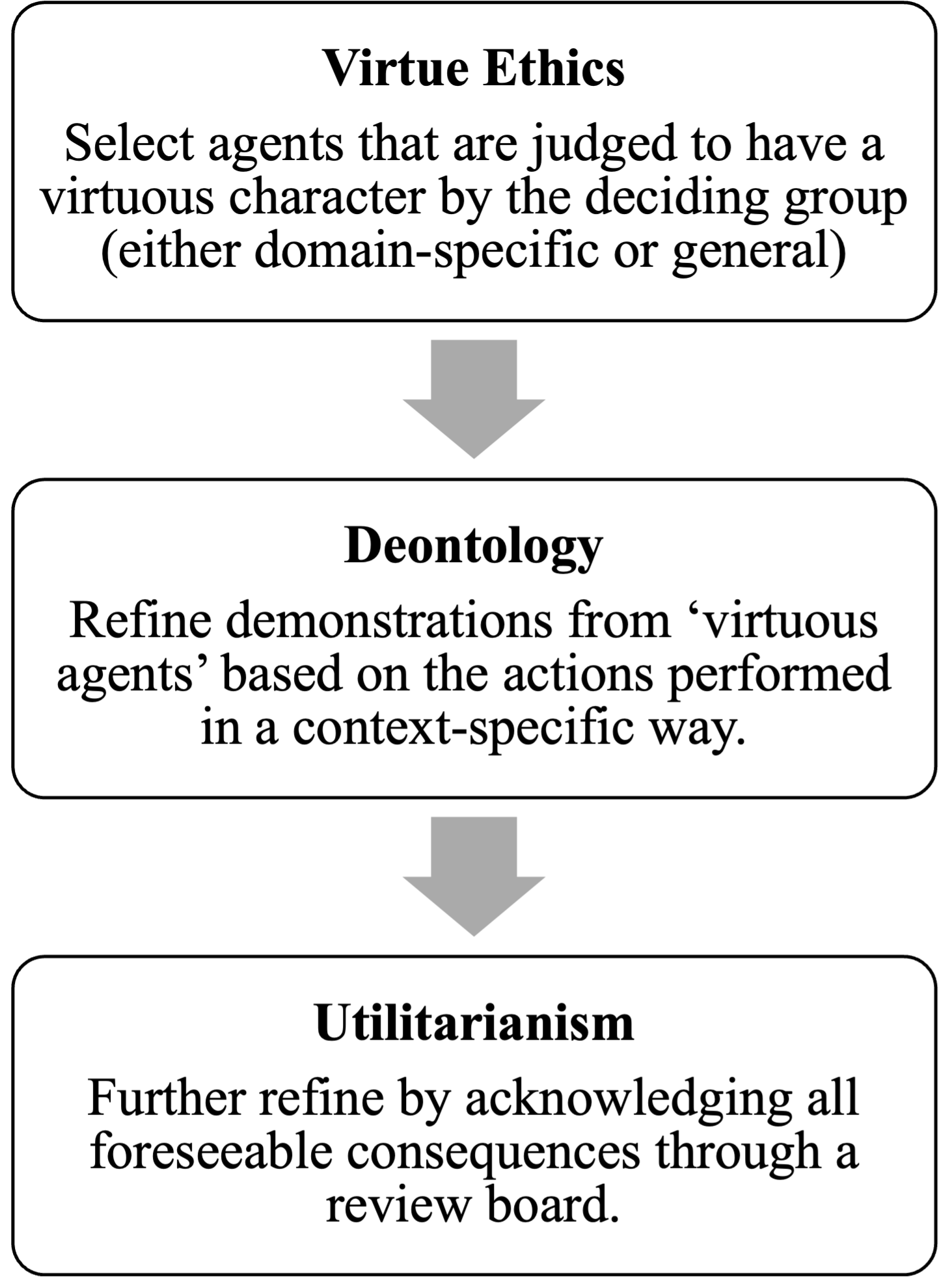}
    \caption{A hierarchical structure could be developed for building ethical AI by combining virtue ethics, deontology and utilitarianism.}
    \label{fig:my_label}
\end{figure}

How do we, as humans, learn our morals and values? Why do they differ so much globally as we have seen with the Moral Machine experiment? Children grow to imitate and follow what they have been bought up to see. It is the complex interaction between nature and nurture that ultimately shapes our values, as is becoming more notable in the field of behavioral epigenetics.

With the advent of apprenticeship learning, i.e. inverse reinforcement learning, it would be reasonable to try and develop similar artificial systems that would learn by viewing the behaviors of humans \textsuperscript{\cite{abel_reinforcement_2016}}. As opposed to reinforcement learning which uses rewards and punishments to learn certain behaviors, inverse reinforcement learning would involve an AI system observing human behaviors and figuring out the goal of the behavior itself. This paves a way for coding the complexity of human ethical values without having to explicitly code every rule for every possible scenario. However, this highlights the need once again for stringent thought into the data AI should learn from. A child growing up with abusive parents would possibly learn that it is right to behave in such a way. An excellent example of how this type of exploitation could affect AI is Tay bot, which learned racial remarks from Twitter trolls \textsuperscript{\cite{neff_talking_2016}}.

Regarding the data an AI should learn from, we could look to the actions of the moral exemplars \textsuperscript{\cite{hindocha_moral_2022}}, and thus to the agent-based ethics approach proposed by Aristotle: virtue ethics. This emphasizes an agent’s moral character in their actions, rather than emphasizing duties (deontology) or consequences (utilitarianism). Aristotle himself stresses that virtue is gained through habit, thus a virtuous person consistently embodies the right actions because they have built the right character. 

For instance, in a similar manner to how we have courts and ordered decision-making processes in place to judge one’s character with a rigorous review panel, it may be possible to have an equivalent scrutiny panel for AI. We could use this as part of a hierarchical structure for building moral AI, in which we could combine choice aspects of different moral paradigms, such as virtue ethics, deontology and utilitarianism, an example of which we provide in (Figure 1). 

Furthermore, tying back to one of the first points regarding the differences in moral ideologies across different cultures, it is important to consider the principles that can be taken from other philosophers, for example, Confucian or Buddhist virtues. The four pillars of Confucian ethics (Yen, love and empathy; Yi, righteousness; Li, veneration and comity; Zhi, wisdom) could also be considered when choosing virtuous people for data collection as the ‘teachers’, or moral exemplars, of the AI.

Hagendorff (2022) performed a reductionist clustering approach to meta-studies on AI ethics guidelines to distill four fundamental AI virtues: Justice, honesty, responsibility and care \textsuperscript{\cite{hagendorff_virtue-based_2022}}. The author suggests that these four virtues are behind the majority of practical AI ethical principles discussed in the literature today, such as fairness, transparency, accountability and social cohesion. Notably, prudence and fortitude were identified as second-order virtues that aim to counteract factors, such as innate bias and peer influence. This provides an important starting point for addressing how we might define and identify moral exemplars for developing ethical AI.

As shown in Figure 1, following the selection of ‘teachers’, or moral exemplars, a refinement process will take place: data will be chosen through the selection of actions that comply with very specific rules imposed that are decided upon a priori, as constituting duties in a deontological fashion. This could be followed by a selection of valid actions which exhibit consequences desired according to utilitarianism or other frameworks, such as being desired by the majority as suggested by Noothigattu et al. (2018) \textsuperscript{\cite{noothigattu_voting-based_2018}}.

It is necessary to highlight an important hypothetical scenario described by Wallach and Allen, regarding AI-based trading systems that lead to overly high oil prices \textsuperscript{\cite{wallach_moral_2009}}. This would lead to other automated systems moving to coal as an alternative source of energy to reduce costs – a massive influx in coal demand and the need for full-time production could lead to an explosion and power outage. As one can see, the channels of multiple automated decision systems in interaction and under the influence of one another could lead to catastrophic human and economic consequences. It is with that, that there has to be a higher-order framework in place to assess all potential interactions between AI systems, for example, the AI influencing the trade of oil with an AI deciding which power source to use.

This ties into the self-awareness embodiment of AI and awareness of other systems as well as humans. Hence, the cognitive simulation theory seems another way to help avoid such scenarios where AI could be programmed through internal simulations of actions with predictions of their consequences, instead of logic statements \textsuperscript{\cite{vanderelst_architecture_2018}}. In this paper, a multi-layer architecture involving a controller that generates several actions to be performed to reach a certain goal is proposed. These actions are then sent to an ethical layer that simulates each action using the current state of the environment, the human involved and the robot. Internal states of the human and robot are retrieved and used in conjunction to evaluate the best action to take. 

With this, it is important to determine the tests and simulations that must be carried out even before AI systems are given higher moral authority. Devin Gonier (2018) introduced the notion of 'tricking' AI models based on the Hawthorne effect - the observer affects the outcome of an experiment \textsuperscript{\cite{devin_gonier_morality_2018}}. Based on this, if we were to constantly observe a model, the outcome should be in our favor. We could have two scenarios. The computer thinks it moves from a simulation to the real world, but in reality, it stays in simulation mode. Conversely, the model could think it's in a simulation but actually in the real world. Ultimately, keeping the model in a 'doubt' mode enables control.

Could we then have parallel systems running that would each take a vote on the best action? If the AI working on trading systems could predict that high oil prices would result in decisions to move away from oil by other logically-programmed machines, it would know to avoid such a scenario. Such a framework could be combined with the inverse reinforcement learning approach where the AI learns from observing actions from exemplars, and figuring out the goals and internal states of impacted agents. In doing so, an extensive learned base could be created to enable internal simulations of actions and accurate predictions of consequences without the need for explicit rules. Such a process would take time and experience to learn, hence the introduction of regulatory sandboxes, which will be discussed in the next section, are a key way to ensure a controlled application of ethical AI.

\section{AI GOVERNANCE PRINCIPLES}

Ultimately, the development of moral machines heavily relies on stringent rules in place for deploying such systems. At the very least, it should be proven that accountability has been covered, the system is transparent for inspection, cannot be easily manipulated (i.e. strong security measures have been employed to reduce the chance of hackers intercepting) and that the AI has a high level of predictability that can match the majority of humans.
Regarding opacity and accountability, it may be possible to have parallel functioning systems, the black box - referring to complex models that are not highly interpretable to humans - and it's more transparent shadow that aims to conduce step-by-step thought processes. Prospector is one such example, whereby researchers have developed an interactive visual representation of predictive models that aims to help us understand how and why data points are predicted, and the predictive value of different features \textsuperscript{\cite{krause_interacting_2016}}. Notably, the authors of this paper stress that we need to move away from solely evaluating performance metrics, such as accuracy.

Research is being done to ensure AI is more transparent, for example, gradient-weighted class activation mapping is one method that attempts to explain convoluted neural networks used in histology or magnetic resonance imaging labeling. Maps are created to demonstrate the most relevant areas of images used in the classification. Such techniques enable histopathologists or radiologists to gain some insight into the decision-making process. Whilst this provides information on how a model made a decision, current explanatory methods typically do not justify whether a decision was appropriate or not. Notably, Ghassemi, Oakden-Rayner and Beam discussed the idea that humans unknowingly use heat maps to understand whether the areas identified are justifiable based on intuition, and thus, bias can be introduced. Humans do tend to over-trust computers, and explainability methods could reduce the ability to identify errors and/or unreasonably increase our confidence in algorithmic decisions \textsuperscript{\cite{ghassemi_false_2021}}.

Indeed, detailed logs must be made available by the AI to help understand any decision and to further guide its development. As suggested by Mitchell et al. (2019), documentation detailing the performance of models in different contexts (e.g. age, race, sex) should be provided \textsuperscript{\cite{mitchell_model_2019}}. Increasing numbers of publications relating to frameworks for dataset generation and AI development are becoming available and a compilation of these academic works must be acknowledged in any regulatory documents going forward.

Incorporating these techniques and tools into the AI development process can help us work towards creating AI systems that reflect the world we want to live in. By fostering transparency, accountability, and fairness, we can ensure that AI systems adhere to the principles of virtue ethics in the age of AI.

Various governments, including the UK, US and Japan, have leaned towards the use of ‘regulatory sandboxes’, defined as a testbed for innovation in a controlled real-life environment prior to launch. The 2023 UK policy paper, "A pro-innovation approach to AI regulation", highlights the use of sandboxes to encourage innovation and provide an adaptable response to regulating AI \textsuperscript{\cite{department_for_pro-innovation_2023}}. This contrasts with proposals for EU legislation (the EU AI Act or the Liability Directive) which take a different approach with more comprehensive regulations, including the General Data Protection Regulation (GDPR) that applies to AI development and data sharing across EU member states \textsuperscript{\cite{european_commission_launch_2022}}. GDPR emphasizes data subject rights, data minimization, and transparency. The UK's Data Protection Act 2018 incorporates the GDPR into UK law, ensuring similar data protection standards for AI development and data sharing. Nevertheless, Spain has piloted the first regulatory sandbox on AI in the EU to aid in generating future guidelines \textsuperscript{\cite{european_commission_launch_2022}}. Many countries have also adopted a sector-specific approach to regulations, impacting AI development and data sharing in various industries, such as finance, energy and healthcare.

As technology constantly evolves, guidelines will also have to evolve with it. There is no set of rules that can foresee all future problems that arise. This is also why it is important to have interdisciplinary research involving engineers, philosophers, neuroscientists, psychologists and policymakers to ensure the best possible outcomes. It may be that regulations require companies to involve a multidisciplinary team during development. University education may also need to introduce varied modules (i.e. ML courses include a philosophy and ethics module, and medical degrees include ML models) to prepare future generations for such collaborative tasks. Regulatory workshops should be made more regular and accessible to inform and guide companies during the development process.
Either way, there is still much yet to debate regarding human ethics before machines can become the moral entities we want them to become. If we are to take the bottom-up approach to train AI, we must ensure the data is a reflection of the world we want to live in.

Encouraging open discussions among individuals and communities with different views will be important in decision-making. This could be done by creating national and international forums, where society can share experiences and develop recommendations for AI governance. Such forums can provide a platform for dialogue, consensus-building, and the exchange of best practices. By fostering a culture of dialogue and debate, society can collectively engage with moral dilemmas to understand and overcome the nuances and complexities. Additionally, individuals can further develop the ability to critically analyze moral dilemmas through the promotion of moral education and critical thinking from an early age, enabling them to understand different viewpoints and appreciate the diversity of moral values across cultures and traditions. We must also ensure that decision-making processes are inclusive and represent the diversity of society. This will help to avoid the domination of a single moral perspective and to embrace multiple moral views.

\section{TOWARDS MORAL AI}

In conclusion, we have looked at the matter of building moral machines from different angles, such as those of technical approaches, culture, emotion, sentience, and governance. Our main contribution is the proposal of a hybrid approach using a combination of top-down and bottom-up and a further combination of different moral paradigms to get around the different obstacles and limitations discussed. We have also discussed some of the practical issues and techniques relevant to implementing our approach. 

Ultimately, determining what constitutes good morals is an ongoing and collaborative effort that requires input from various stakeholders and an openness to learn, grow, and adapt. By engaging in these processes, society can work towards establishing shared moral values that contribute to the greater good. It is in doing so, that we may be able to start to answer some of the questions we, as humans, hold regarding morality and what it is to act with moral grounding. This collaborative approach can help create a more responsible, inclusive, and beneficial AI landscape for all.

\section{ACKNOWLEDGMENTS}
Reneira Seeamber was supported by the UKRI CDT in AI for Healthcare http://ai4health.io (Grant No. EP/S023283/1). For the purpose of open access, the authors have applied a Creative Commons Attribution (CC BY) license to the Accepted Manuscript.

\def\refname{REFERENCES}

\vspace*{-8pt}

\end{document}